\documentclass[11pt]{article}
\usepackage{coling2020}
\usepackage{times}
\usepackage{url}
\usepackage{latexsym}
\usepackage{xcolor}
\usepackage{graphicx}
\usepackage{amsmath}
\usepackage{subfigure}
\usepackage{multirow}
\usepackage{booktabs}
\usepackage{microtype}
\usepackage{stfloats}
\usepackage{amsfonts,amssymb}
\usepackage{bbm}
\usepackage{hyperref}
\usepackage{caption}

\colingfinalcopy 

\title{Infusing Sequential Information into Conditional Masked Translation Model with Self-Review Mechanism}

\author{Pan Xie\thanks{\enspace The work was done as an intern at Xiaomi.} $^1$ \enspace Zhi Cui$^2$ \enspace Xiuying Chen$^3$ \enspace Xiaohui Hu$^4$ \enspace Jianwei Cui$^2$ \enspace Bin Wang$^2$ \\ $^1$Beihang University \\ $^2$Xiaomi AI Lab \\ $^3$Peking University \\ $^4$Chinese Academy of Sciences \\ 
{\tt \small panxie@buaa.edu.cn \enspace xy-chen@pku.edu.cn \enspace hxh@iscas.ac.cn} \\ 
\tt \small \{cuizhi,cuijianwei,wangbin11\}@xiaomi.com}
  
\date{123}

\begin{document}
\maketitle
\begin{abstract}
Non-autoregressive models generate target words in a parallel way, which achieve a faster decoding speed but at the sacrifice of translation accuracy. To remedy a flawed translation by non-autoregressive models, a promising approach is to train a conditional masked translation model (CMTM), and refine the generated results within several iterations. Unfortunately, such approach hardly considers the \textit{sequential dependency} among target words, which inevitably results in a translation degradation. Hence, instead of solely training a Transformer-based CMTM, we propose a Self-Review Mechanism to infuse sequential information into it. Concretely, we insert a left-to-right mask to the same decoder of CMTM, and then induce it to autoregressively review whether each generated word from CMTM is supposed to be replaced or kept. The experimental results (WMT14 En$\leftrightarrow$De and WMT16 En$\leftrightarrow$Ro) demonstrate that our model uses dramatically less training computations than the typical CMTM, as well as outperforms several state-of-the-art non-autoregressive models by over 1 BLEU. Through knowledge distillation, our model even surpasses a typical left-to-right Transformer model, while significantly speeding up decoding. 
\end{abstract}

\section{Introduction}
Neural Machine Translation (NMT)  models have achieved a great success in recent years~\cite{Sutskever2014SequenceTS,Bahdanau2015NeuralMT,Cho2014LearningPR,Kalchbrenner2016NeuralMT,Gehring2017ConvolutionalST,Vaswani2017AttentionIA}. Typically, NMTs use autoregressive decoders, where the words are generated one-by-one. However, due to the left-to-right dependency, this computationally-intensive decoding process cannot be easily parallelized, and therefore causes a large latency~\cite{Gu2018NonAutoregressiveNM}. 

To break the bottleneck of autoregression, several non-autoregressive models have been proposed to induce the decoder to generate all target words simultaneously~\cite{Gu2018NonAutoregressiveNM,Kaiser2018FastDI,Li2019HintBasedTF,Ma2019FlowSeqNC}. Despite the acceleration of computation efficiency, these models usually suffers from the cost of translation accuracy. Even worse, they decode a target only in one shot, and thus miss a chance to remedy a flawed translation. Against them, a promising research line is to perform refinement on the generated result within several iterations~\cite{lee2018deterministic,ghazvininejad2019mask}.

Along this line, \newcite{ghazvininejad2019mask} propose a Mask-Predict decoding strategy, which iteratively refines the generated translation given the most confident target words predicted from the previous iteration. This model is trained using an objective of conditional masked translation modeling (CMTM), by predicting the masked words conditioned on the rest of observed words. However, CMTM just learns from a subset of words instead of the entire target in terms of a training step. As a result, it will iterate more times over the training dataset to explore the contextual relationship within a sentence, and thus will struggle with a huge cost of the whole training time~\cite{Clark2020ELECTRAPT}. Most importantly, CMTM extensively bases upon the assumption of conditional independence, making it hard to capture the strong correlation across the adjacent words~\cite{Gu2018NonAutoregressiveNM}. Inevitably, this issue will still degrade the translation performance, such as outputting repetitive words~\cite{Wang2019NonAutoregressiveMT}.

\begin{figure}[tp]
\centering
\includegraphics[width=0.7\textwidth]{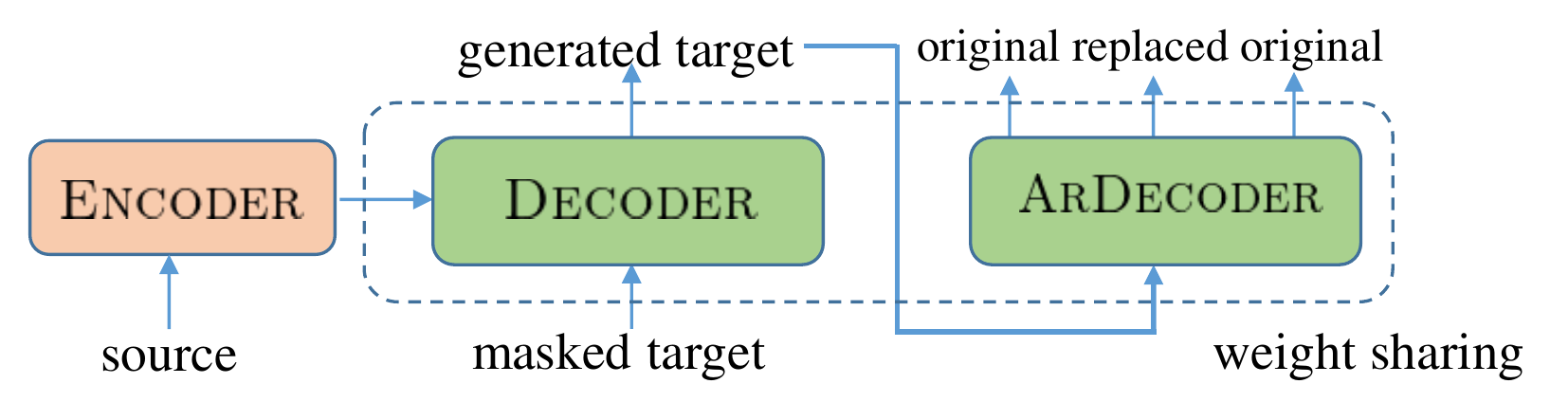}
\caption{A simplified architecture of our model based on Transformer. We tie the weights of \textsc{Decoder} and \textsc{ArDecoder}. The inter attention between \textsc{Encoder} and \textsc{ArDecoder} is omitted for readability.}
\label{simple-arch} 
\end{figure}

To address the issues, our idea is to infuse sequential information into CMTM. Accordingly, we propose a Self-Review Mechanism, a discriminative task in which the model learns to autoregressively distinguish the ground truth target from the non-autoregressively generated output of itself. As shown by \textsc {ArDecoder} (short for Autoregressive Decoder) in Figure~\ref{simple-arch}, we firstly switch on the autoregressive mode of the same \textsc{Decoder} with CMTM by inserting a left-to-right mask. More importantly, we then require this \textsc{ArDecoder} to recurrently review whether each generated word from CMTM is supposed to be a replacement or just an original. In this way, this mechanism constrains our model to review each predicted word only based on previous ones, which is able to not only correct the prediction errors, but also facilitates the learning of conditional dependence of the target words. Moreover, this mechanism could also help speed up the whole training by learning from all target words rather than a small masked subset.

We extensively validate our model on the datasets of WMT14 En$\leftrightarrow$De and WMT16 En$\leftrightarrow$Ro. The experimental results demonstrate that our model outperforms several state-of-the-art non-autoregressive models by over 1 BLEU. Through knowledge distillation, our model even achieves competitive performance compared with the typical left-to-right Transformer, while significantly reducing the cost of time during inference. Meanwhile, we also prove that the training speed of our model is much faster than the typical CMLM.

\section{Background}

\subsection{Autoregressive NMT}
Given a source sentence \(\mathbf{x}=\{x_1,x_2,...,x_{|\mathbf{x}|}\}\), a NMT model is aimed to generate a sentence in target language \(\mathbf{y}=\{y_1,y_2,...,y_{|\mathbf{y}|}\}\) with identical semantics expressed, where $|\mathbf{x}|$ and $|\mathbf{y}|$ are denoted as the length of source and target sentence, respectively. Typically, the training objective of an autogressive NMT model is expressed as a chain of conditional probabilities in a left-to-right manner:

\begin{equation}
\begin{aligned}
\log{p_{at}(\mathbf{y}|\mathbf{x})} = \sum_{t=1}^{|\mathbf{y}|+1}\log{p(y_t|y_{0:t-1}, \mathbf{x})}
\end{aligned}
\label{eqn:equation1}
\end{equation}

\noindent where $y_0$ and $y_{|\mathbf{y}|+1}$ are $<$SOS$>$ and $<$EOS$>$, standing for the start and end of a sentence, respectively. Usually, these probabilities are parameterized using a standard encoder-decoder architecture~\cite{Sutskever2014SequenceTS}, where the decoders use autoregressive strategy to capture the left-to-right dependency among the target words.

\subsection{Conditional Masked Translation Model}
Different from the training objective in Equation~\ref{eqn:equation1}, we adopt conditional masked translation modeling (CMTM)~\cite{ghazvininejad2019mask} to optimize our proposed non-autoregressive NMT model. During training, our model is aimed to predict a set of masked target words $\mathbf{y}_{mask}$ given an source input $\mathbf{x}$ and a set of observed target words $\mathbf{y}_{obs}$. Note that $|\mathbf{y}| = |\mathbf{y}_{mask}| + |\mathbf{y}_{obs}|$. Based on the assumption that the words of $\mathbf{y}_{mask}$ are independent, the training objective of CMTM is formulated as:

\begin{equation}
\begin{aligned}
\log{p_{nat}(\mathbf{y}_{mask}|\mathbf{x}, \mathbf{y}_{obs})} = \sum_{t=1}^{|\mathbf{y}_{mask}|}\log{p(y^{t}_{mask}|\mathbf{x}, \mathbf{y}_{obs})}
\end{aligned}
\label{eqn:equation2}
\end{equation}

\noindent where the masked words in $\mathbf{y}_{mask}$ are randomly selected and denoted by a special token $[$mask$]$.  

\section{Approach}
\subsection{Model Architecture}
\begin{figure*}[t]
\centering
\includegraphics[width=0.9\textwidth]{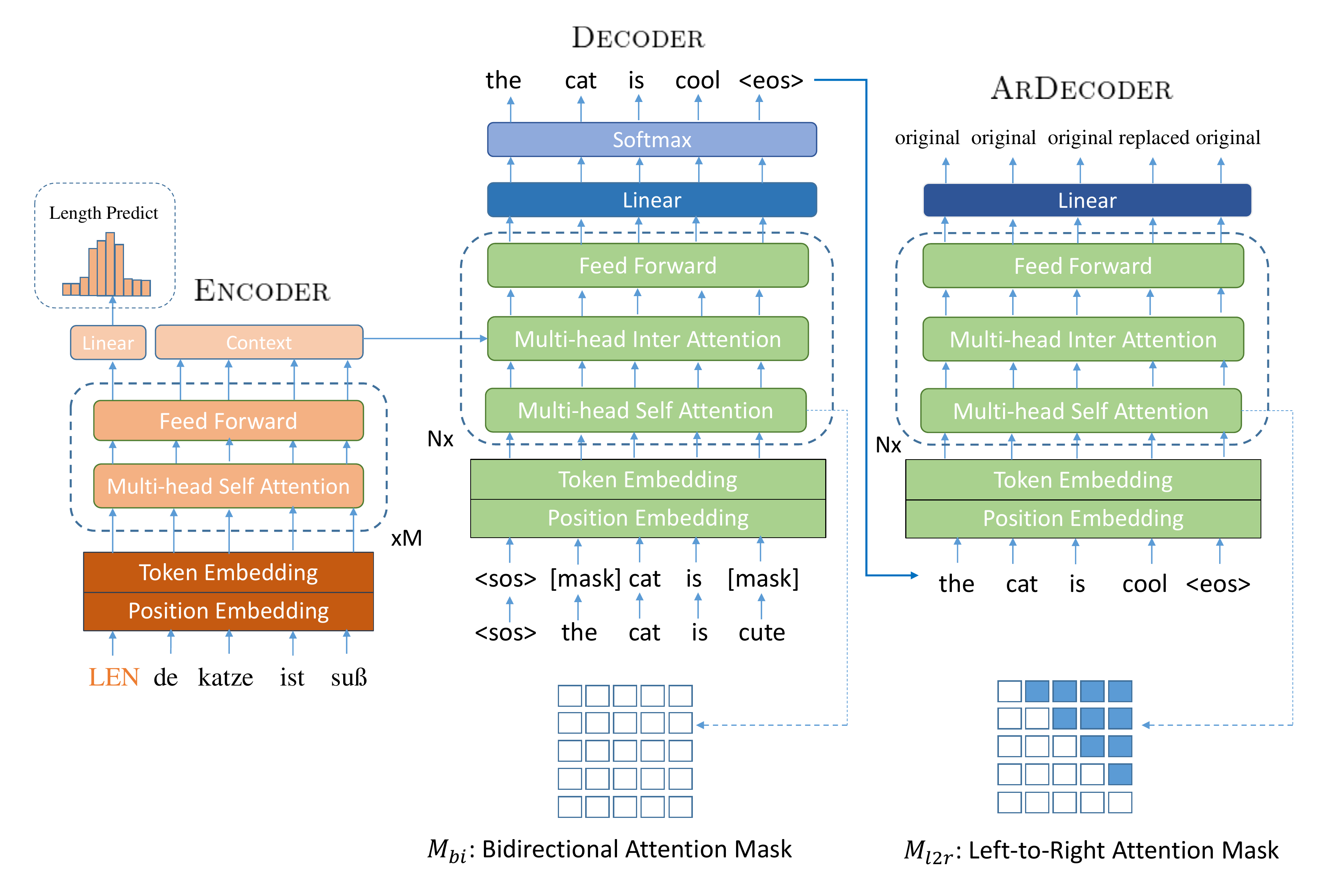}
\caption{The architecture of our model, which includes an \textsc{encoder}, a \textsc{decoder} and a \textsc{ArDecoder}. During inference, we only perform iterative decoding based on \textsc{Encoder-Decoder}. The weights are shared across the \textsc{Decoder} and \textsc{ArDecoder} as denoted by the green blocks.}
\label{architecture} 
\end{figure*}

Figure~\ref{architecture} illustrates the overall architecture of our proposed model, which is composed of three modules, an \textsc{Encoder}, a \textsc{Decoder} and an \textsc{ArDecoder}. Notably, \textsc{ArDecoder} is obtained by solely adding a left-to-right mask to \textsc{Decoder}, where their weights are tied. Rather than a pure CMTM, we also propose a Self-Review Mechanism to ask the \textsc{ArDecoder} to review the predicted target from \textsc{Decoder} in a left-to-right manner. In this section, we will detail each module and the Self-Review Mechanism.

\paragraph{Encoder} Our \textsc{Encoder} is identical to the standard Transformer \cite{Vaswani2017AttentionIA}. Built upon self attention, it encodes a source input $\mathbf{x}$ into a series of contextual representations \(\mathbf{H}_{enc}=\{h^1_{enc}, h^2_{enc}, ..., h^{|\mathbf{x}|}_{enc}\}\) by:

\begin{equation}
\begin{aligned}
\mathbf{H}_{enc}=\{h^1_{enc}, h^2_{enc}, ..., h^{|\mathbf{x}|}_{enc}\} = \text{encode}(\mathbf{x})
\end{aligned}
\label{eqn:equation3}
\end{equation}

\paragraph{Decoder} The non-autoregessive property of our model mainly lies in our \textsc{Decoder}. Different from \textsc{Encoder}, the \textsc{Decoder} has two sets of attention heads as shown in Figure~\ref{architecture}: the inner heads are attending over the target words, and the inter heads are over the hidden outputs of \textsc{Encoder}. It is worth noting that we use a bidirectional mask  (denoted as $M_{bi}$) as shown in the middle of Figure~\ref{architecture}. Such mask allows the \textsc{Decoder} to use both left and right contexts to predict each target word, ensuring that the prediction for $t$-th position can depend not only on the information before $t$-th but also right after $t$-th.

Our \textsc{Decoder} is optimized using the objective in Equation~\ref{eqn:equation2}. Given a source input $\mathbf{x}$ and part of observed target words $\mathbf{y}_{obs}$, the \textsc{Decoder} is required to predict those words of $\mathbf{y}_{mask}$. Firstly, we obtain a series of \textsc{Decoder} hidden outputs $\mathbf{H}_{dec}=\{h^1_{dec}, h^2_{dec},...,h^{|\mathbf{y}|}_{dec}\}$, by feeding \textsc{Decoder} with the observed target $\mathbf{y}_{obs}$ and the \textsc{Encoder} outputs $\mathbf{H}_{enc}$. Mathematically, we parameterize $\mathbf{H}_{dec}$ as:

\begin{equation}
\begin{aligned}
\mathbf{H}_{dec}=\{h^1_{dec}, h^2_{dec},...,h^{|\mathbf{y}|}_{dec}\} = \text{decode}(\mathbf{H}_{enc}, \mathbf{y}_{obs}; M_{bi})
\end{aligned}
\label{eqn:equation4}
\end{equation}

Then, we apply a linear projection on the hidden outputs $\mathbf{H}_{dec}$, and obtain the probabilities of target words using softmax. Notably, we only focus on the probabilities of masked words during training. Therefore, the probability $p(y^{t}_{mask}|\mathbf{x}, \mathbf{y}_{obs})$ in Equation~\ref{eqn:equation2} is parameterized by: 

\begin{equation}
\begin{aligned}
p(y^{t}_{mask}|\mathbf{x}, \mathbf{y}_{obs}) = \text{softmax}(W_1^Th^t_{dec})
\end{aligned}
\label{eqn:equation5}
\end{equation}


\paragraph{\textsc{ArDecoder}} The \textsc{ArDecoder} is introduced to serve as a discriminator of adversarial models~\cite{goodfellow2014generative}, and will play an important role in our Self-Review Mechanism. As shown in Figure~\ref{architecture}, \textsc{ArDecoder} is obtained by adding a left-to-right mask (denoted as $M_{l2r}$) to the \textsc{Decoder}. This mask prevents \textsc{ArDecoder} from attending future words when reading the predicted target from \textsc{Decoder}, ensuring that the prediction for $t$-th position can only rely on the known outputs before $t$-th. 

Unlike the aim of \textsc{Decoder}, \textsc{ArDecoder} is asked to review the predicted sentence from \textsc{Decoder}, and distinguish whether each word is supposed to be replaced or not. Notably, we tie the weights of \textsc{ArDecoder} and \textsc{Decoder}, to ensure that our \textsc{Decoder} can take advantage of the sequential information learned from this discriminative task. 

\subsection{Self-Review Mechanism}
As discussed previously, solely a CMTM is insufficient to capture the sequential dependency of target words, and thus still inevitably results in a disappointing translation. To remedy the issue, the core of our work is how to better \textit{infuse} sequential information into the model.

Concretely, we propose a Self-Review Mechanism for CMTM to learn strong correlations among the target words. During training, \textsc{Encoder-Decoder} firstly predicts a target $\mathbf{\hat{y}}$ given a gold observed target $\mathbf{y}_{obs}$ as well as an input source $\mathbf{x}$. Then, the \textsc{ArDecoder} is asked to review  the predicted target $\mathbf{\hat{y}}$, and distinguish whether a predicted word $\hat{y_t}$ is supposed to be replaced by the ground truth $y_t$ as:

\begin{equation}
\begin{aligned}
p(\hat{y_t}=y_t |\hat{y}_{1:t-1}, \mathbf{x}) = \sigma(W_2^Th^t_{rev})
\end{aligned}
\label{eqn:equation6}
\end{equation}

\begin{equation}
\begin{aligned}
\mathbf{H}_{rev} = \{ h^1_{rev}, h^2_{rev}, ..., h^{|\mathbf{y}|}_{rev} \} = \text{review}(\mathbf{H}_{enc}, \mathbf{\hat{y}}; M_{l2r})
\end{aligned}
\label{eqn:equation7}
\end{equation}


\noindent where $\sigma(\cdot)$ is a sigmoid function. Finally, the objective of the self-reviewing becomes: 

\begin{equation}
\begin{aligned}
\mathcal{L}_{rev} = -\sum_{t=1}^{|\mathbf{y}|}( (\hat{y_t}={y_t})\log{(\sigma(W_2^Th^t_{rev}))} + (\hat{y_t} \neq{y_t})\log{(1- \sigma(W_2^Th^t_{rev}))}   )
\end{aligned}
\label{eqn:equation8}
\end{equation}

Here, we do not back-propagate the learning errors from \textsc{ArDecoder} to \textsc{Encoder-Decoder} due to the difficulty of applying adversarial learning to text~\cite{Caccia2020LanguageGF}. By adding up $\mathcal{L}_{rev}$, our model sees the entire target sentence rather than a small subset of words in terms of a training step, and thus it does not need to iterate more times to explore the contextual relationships among the words, which is beneficial to speeding up the whole training compared with a pure CMTM~\cite{Clark2020ELECTRAPT}. Besides, our work can also be regarded as a multi-task learning, where we enhance our \textsc{Decoder} with the bidirectional contextual information as well as the left-to-right correlations of target words.

\subsection{Length Prediction}
Typically, an autoregressive NMT model generates the target sentence word-by-word, and thus it decides the length of target sentence by encountering a special token $<$EOS$>$. However, our model adopts the strategy of non-autoregressive decoding, namely, it predicts the entire target sentence in a parallel way. Following \cite{Devlin2019BERTPO,ghazvininejad2019mask}, we add a special token $<$LEN$>$ to the begining of source input. In this sense, our \textsc{Encoder} is also required to predict the length of target sentence $L$, i.e., predict a token from $[1, N]$ given the source input $\mathbf{x}$, where $N$ is the maximum length of target sentences in our corpus. Mathematically, we define the loss of length prediction as:

\begin{equation}
\begin{aligned}
\mathcal{L}_{len} = \sum_{i}^N-(L=i)\log{p(L|\mathbf{x})}
\end{aligned}
\label{eqn:equation9}
\end{equation}

\subsection{Optimization and Inference}
Overall, the whole model is jointly trained by minimizing the total loss $\mathcal{L}$, which is a combination of Equation~\ref{eqn:equation2},\ref{eqn:equation8},\ref{eqn:equation9}:

\begin{equation}
\begin{aligned}
\mathcal{L} = \mathcal{L}_{dec} + \mathcal{L}_{len} + \mathcal{L}_{rev} 
\end{aligned}
\label{eqn:equation10}
\end{equation}

\noindent where $\mathcal{L}_{dec}=-\log{p_{nat}(\mathbf{y}_{mask}|\mathbf{x}, \mathbf{y}_{obs})}$. 

During inference, we abandon \textsc{ArDecoder} and perform iterative refinement only based on \textsc{Encoder-Decoder}. Following Mask-Predict~\cite{ghazvininejad2019mask}, we generate a raw sequence starting with an entirely masked target given a new input source. Upon this raw sequence, we conduct refinement by masking-out and re-predicting a subset of words whose probabilities are under a threshold. This refinement is repeated within a heuristic number of iterations. For more details, please refer to \cite{ghazvininejad2019mask}.

\section{Experiments}
\subsection{Setting}
\paragraph{Datasets}We conduct experiments on two benchmark datasets, WMT14 En\(\leftrightarrow\)De (4.5M sentence pairs) and WMT16 En\(\leftrightarrow\)Ro (610k pairs). After preprocessing the two datasets following~\cite{lee2018deterministic}, we tokenize them into subword units using BPE~\cite{sennrich2016neural}.  We use newstest-2013 and newstest-2014 as our development and test datasets for WMT14  En\(\leftrightarrow\)De, while use newsdev-2016 and newstest-2016 as our development and test datasets for WMT16 En\(\leftrightarrow\)Ro.

\paragraph{Evaluation Metrics} We adopt the widely-used \textbf{BLEU}\footnote{https://github.com/moses-smt/mosesdecoder/blob/master/scripts/generic/multi-bleu.perl}~\cite{Papineni2001BleuAM} to evaluate the translation accuracy. To compare the training speed, we also use \textbf{Floating-Point Operations per second} (FLOPs)\footnote{https://github.com/google-research/electra/blob/master/flops\_computation.py} to measure the computational complexity.

\paragraph{Implementation Details} We follow the base configuration of Transformer~\cite{Vaswani2017AttentionIA}: The dimension of model is set to 512, and the dimension of inner layers is set to 2048. The \textsc{Encoder} is consisted of a stack of 6 layers , as well as the \textsc{Decoder} and \textsc{ArDecoder}. The weights of our model are all randomly initialized with a uniform distribution \(\mathcal N(0, 0.02)\). Besides, we set the parameters of layer normalization as \(\beta=0, \gamma=1\). We use Adam optimizer~\cite{Kingma2015AdamAM} with 98k tokens per batch. We increase the learning rate from 0 to 5e-4 within the first 10,000 warmup steps, and gradually decay it with respect to the inverse square root of training steps. Note that we share the weights of \textsc{Decoder} and \textsc{ArDecoder} only except the output layer ($W_1\neq{W_2}$ in Equation~\ref{eqn:equation5} and \ref{eqn:equation6}). During inference, we set length candidates as 5 for non-autoregressive decoding, where the max length $N$ is defined as 10,000. The number of iteration for refinement is set as 10. To compare with autoregressive models, we adopt a beam width of 5 for beam search decoding. The training speed is measured on 8 NVIDIA Tesla P100 GPUs and decoding speed is just on one.

\paragraph{Knowledge Distillation} Previous works on non-autoregressive NMT models have proved that knowledge distillation can substantially improve the performance~\cite{Gu2018NonAutoregressiveNM,lee2018deterministic,Stern2019InsertionTF,Zhou2020UnderstandingKD}. Commonly, a student model is trained on a distilled dataset which is generated by a teacher model, where the teacher model usually adopts a much larger configuration of parameters than its student. Different from this common setting, we will investigate if it is still useful to tie the configuration of the teacher and its student model. We train our model on a distilled corpus (EN\(\leftrightarrow\)DE and EN\(\leftrightarrow\)RO), where the distilled target are generated by a typcial left-to-right Transformer with a base configuration. In the followings, we will identify the effect of knowledge distillation to our model.

\subsection{Baselines} To demonstrate the effectiveness of our work, we compare with several state-of-the-art NMT models:

\noindent \textbf{Seq2Seq}~\cite{Bahdanau2015NeuralMT}: It is a LSTM-based sequence-to-sequence model, where the decoder adopts beam search strategy. 

\noindent \textbf{ConvS2S}~\cite{Gehring2017ConvolutionalST}: It is a  convolution-based sequence-to-sequence model, and it decodes the target words in a left-to-right manner.

\noindent \textbf{Transformer}~\cite{Vaswani2017AttentionIA}: It is a state-of-the-art autoregressive model, and it adopts beam search decoding to generate target translation.

\noindent \textbf{FTNAT}~\cite{Gu2018NonAutoregressiveNM}: It is a non-autogressive Transformer model using fertitilies, and adopts noisy parallel decoding (NPD) to generate target translation. 

\noindent \textbf{FlowSeq}~\cite{Ma2019FlowSeqNC}: It is also a non-autogressive model, which introduces a latent variable to model the generative flow. During inference, it generates a target translation using argmax decoding.

\noindent \textbf{HintNAT}~\cite{Li2019HintBasedTF}: It is also a non-autoregressive model, which leverags alignments and hidden states of a teacher autoregressive model.

\noindent \textbf{IRNAT}~\cite{lee2018deterministic}: It is a non-autogressive model trained with a conditional denoising autoencoder. During inference, it iteratively devises the generated translation. We set the number of iterations as 10.

\noindent \textbf{Mask-Predict}~\cite{ghazvininejad2019mask}: It is a typical CMTM model. During inference, it adopts  Mask-Predict on the translation within 10 iterations. By comparing with it comprehensively, we aim to examine the effectiveness of our proposed Self-Review Mechanism.

\begin{table*}[h!]
\centering
\smallskip
\resizebox{1.0\textwidth}{!}{
\begin{tabular}{lcccccc}
\toprule[1pt]
\multirow{2}{*}{\textbf{Models}} & \multirow{2}{*}{\textbf{Iterations}} & \multicolumn{2}{c}{\textbf {WMT'14}} & \multicolumn{2}{c}{\textbf {WMT'16}} & \multirow{2}{*}{\textbf{Speedup}}  \\
&  & \textbf {EN\(\rightarrow\)DE} & \textbf {DE\(\rightarrow\)EN} & \textbf {EN\(\rightarrow\)RO} & \textbf {RO\(\rightarrow\)EN} \\
\hline
\textit{\textbf{Autoregressive Models}} & & & & & & \\
\hline
LSTM Seq2Seq & \small{N} &  \small{24.60} & \small{-} & \small{-} & \small{-} & \small{-} \\
ConvS2S & \small{N} & \small{26.42} & \small{-} & \small{-} & \small{-} & \small{-} \\
Transformer & \small{N} &  \small{\underline{27.30}} & \small{\underline{31.09}} & \small{\underline{34.28}} & \small{ \underline{33.99}} & \small{1.00x} \\
\hline
\textit{\textbf{Non-Autoregressive Models}} & & & & & & 
 \\
\hline
FTNAT(w/ kd) & \small{1} & \small{19.17} & \small{23.20} & \small{29.79} & \small{31.44} & \small{2.36x} \\
HintNAT(w/ kd) & \small{1} & \small{25.20} & \small{29.52} & \small{-} & \small{-} & \small{17.80x} \\
FlowSeq-large (w/ kd) & \small{1} & \small{23.72} & \small{28.39} & \small{-} & \small{29.73}  & \small{-} \\
IRNAT(w/ kd) & \small{1} & \small{13.91} & \small{16.77} & \small{24.45} & \small{25.73} & \small{-} \\
& \small{10} & \small{21.61} & \small{25.48} & \small{29.32} & \small{30.19} & \small{5.78x} \\
Mask-Predict(w/ kd) & \small{1} & \small{18.05} & \small{21.83} & \small{27.32} & \small{28.20} & \small{27.51x} \\
& \small{4} &  \small{25.94} & \small{29.90} & \small{32.53} & \small{33.23} & \small{11.21x} \\
& \small{10} & \small{\underline{27.03}} & \small{\underline{30.53}} & \small{\underline{33.08}} &  \small{\underline{33.31}} & \small{5.16x} \\
\hline
\textbf{Ours(w/o kd)} & \small{1} & \small{9.01} & \small{8.76} & \small{20.76} & \small{20.25} & \small{27.51x} \\
& \small{4} & \small{23.33} & \small{26.58} & \small{33.08} & \small{33.26} & \small{11.21x} \\
& \small{10} & \small{25.50} & \small{29.27} & \small{\textbf{34.54}} & \small{\textbf{34.36}} & \small{5.16x} \\
\hline
\textbf{Ours(w/ kd)} & \small{1} & \small{17.48} & \small{21.47} & \small{23.03} & \small{24.30} & \small{27.51x} \\
& \small{4} & \small{26.99} & \small{30.67} & \small{32.83} & \small{33.55} & \small{11.21x} \\
& \small{10} & \small{\textbf{27.97}} & \small{\textbf{31.59}} & \small{33.98} & \small{\textbf{34.34}} & \small{5.16x}  \\
\bottomrule[1pt]
\end{tabular}
}
\caption{The BLEU scores of all models on the benchmark datasets, where ``kd" is denoted as knowledge distillation. In the column of speedup, we adopt seconds/sentence to measure the decoding speed, where Transformer is set as the baseline (beam size = 5). We present the best BLEU scores of the baseline models reported in their original paper.}
\label{main results table}
\end{table*}

\subsection{Comparison Against Baselines}
The experimental results are summarized in Table~\ref{main results table}. We firstly examine the non-autoregressive models with different decoding strategies, i.e., one-shot decoding vs iterative decoding. As shown in Table~\ref{main results table}, FTNAT, HintNAT and FlowSeq achieve the lowest score of BLEU. Such degradation is mainly due the problem of multimodality~\cite{Gu2018NonAutoregressiveNM} that these models hardly considers the left-to-right dependency. Even worse, they do not have a chance to remedy the translations. The same thing happens to the first iteration of IRNAT, Mask-Predict and our model as well, where the results are similar to the one-shot decoding models. From this comparison, we can conclude that iterative decoding is an effective technique for non-autoregressive NMTs. 

Although IRNAT and MaskPredict are able to turn the initial bad translation into a much better one through multiple iterations of decoding, there is still a gap of the translation accuracy when comparing against the SOTA autoregressive model, i.e., Transformer. Still, this deficiency is attributed to the lack of a mechanism or strategy to capture the strong correlations among the target words, which is also the root cause why non-autoregressive models are hard to generated satisfactory translation~\cite{ren2020study}.

In contrast, our model, which is additionally optimized with our proposed Self-Review Mechanism, significantly achieves a performance boost over these non-autoregressive models.  Meanwhile, our model has a huge lead in BLEU on the dataset of WMT 14 EN\(\rightarrow\)DE compared with Seq2Seq and ConvS2s, and even accomplishes comparable performance with Transformer. More specifically, compared with Transformer, our model (w/o kd) achieves 34.54 (+0.26 gains) and 34.36 (+0.37 gains) of BLEU  on WMT16 EN\(\rightarrow\)RO and WMT16 RO\(\rightarrow\)EN, respectively. Even with the help of knowledge distillation, our model outperforms Transformer on almost all the benchmark datasets except WMT16 EN\(\rightarrow\)RO. More importantly, our model dramatically reduces the cost in decoding, which is at least 5.16x faster than Transformer. If we sacrifice a certain translation accuracy by reducing iteration number, we could obtain even higher decoding efficiency. In brief, this comparison results validate the effectiveness of Self-Review Mechanism.

\subsection{Effect of Knowledge Distillation}
\label{Effect of Knowledge Distillation}
The comparison results are listed in the last 6 rows of Table~\ref{main results table}. In terms of the large-scale dataset, i.e., WMT14 EN$\leftrightarrow$De, our model with the knowledge distillation gains a remarkable improvement, especially at the early iterations. Under the same size of configuration, it is widely believed that that the autoregressive model is better that capturing the alignment relationship between a source-target pair~\cite{Gu2018NonAutoregressiveNM}, and thus the autoregressive model as a teacher model is able to reduce the redundant and irrelevant alignment ``modes" in the raw corpus. In this way, our proposed model benefits from learning such kind of distilled dataset. However, the improvement is not concurrent on the small-scale dataset, i.e., WMT16 EN$\leftrightarrow$RO. At the end of 10th iteration, our model even has a decrease of BLEU on WMT16 EN$\leftrightarrow$RO. We conjecture that a small-scale dataset is statistically likely to contain less redundant ``modes" than a large-scale dataset. As a result, distillation for a small-scale dataset might not be more beneficial for a student model compared with the a raw dataset, probably no matter how large the teacher model is. Therefore, it is useful and more efficient to adopt a teacher model with the same size of configuration as the student model for knowledge distillation on a large-scale dataset.

\section{Ablation Study and Analysis}
Upon CMLM, we additionally introduce a Self-Review Mechanism during training, whereas Mask-Predict~\cite{ghazvininejad2019mask} is optimized with only the first two terms in Equation~\ref{eqn:equation10}. During inference, we abandon \textsc{ArDecoder}, and our model performs decoding as same as Mask-Predict. In this section, we will compare closely to  Mask-Predict to validate the contribution of our proposed Self-Review Mechanism.

\subsection{Training Speed}
To better understand the comparison of training speed between Mask-Predict and our model, we measure the FLOPS of one single step and the whole training steps as shown in Table~\ref{flops} and Table~\ref{flops_compare}, respectively. In terms of the time usage of one training step, Table~\ref{flops} shows that Mask-Predict is about 1.6x faster than our model, since our model has to  optimize \textsc{ArDecoder} together. However, such result of one training step cannot lead to a conclusion that it will take more time to train our model than Mask-Predict. Instead, the results from Table~\ref{flops_compare} illustrate that our model effectively speeds up the whole training especially on a large-scale dataset WMT14 EN\(\leftrightarrow\)DE (at least 5x faster). This discrepancy between one step and overall steps might be resulted from several reasons. We conjecture that our model is able to see whole target sentence, where the \textsc{ArDecoder} needs to review each word generated from \textsc{Decoder}. On the contrary, Mask-Predict only learns from a subset of masked words, and thus it has to take much more steps to discover the semantic relationships among the words. 

\begin{table}[h!]
\centering
\resizebox{0.5\textwidth}{0.04\textheight}{
\begin{tabular}{cccc}
\hline
\textbf{Module} & \textbf{Encoder} & \textbf{Decoder} & \textbf{Reviewer}  \\
\textbf{FLOPs} & 19332M & 24017M & 19468M \\
\hline
\textbf{Model} & \textbf{MaskPredict} & \textbf{Ours} \\
\textbf{FLOPs}  &  43349M & 72484M \\
\hline
\end{tabular}}
\caption{The comparison of FLOPs per training step between Mask-Predict and Ours, where MaskPredict is just composed of \textsc{Encoder-Decoder}, while Ours is composed of \textsc{Encoder-Decoder-ArDecoder}.}
\label{flops}
\end{table}

\begin{table*}[h!]
\centering
\resizebox{\textwidth}{!}{
\begin{tabular}{ccccc}
\toprule[1pt]
\textbf{Dataset/} & \textbf{WMT14 EN\(\rightarrow\)DE} & \textbf{WMT14 DE\(\rightarrow\)EN} & \textbf{WMT16 EN\(\rightarrow\)RO} &\textbf{WMT16 RO\(\rightarrow\)EN} \\
\textbf{Model} &  FLOPs(speedup)  &  FLOPs(speedup)  &  FLOPs(speedup) & FLOPs(speedup)  \\
\hline
\textbf{MaskPredict} (w/ kd)  & 1.12e19 (1.00x) & 6.84e18 (1.00x) &1.33e18 (1.00x) & 1.45e18 (1.00x) \\
\textbf{Ours} (w/ kd) & 1.43e18(\textbf{7.83x}) & 1.30e18 (\textbf{5.26x}) & 0.78e18 (\textbf{1.71x}) & 0.79e18 (\textbf{1.83x}) \\
\bottomrule[1pt]
\end{tabular}}
\caption{Comparison of overall training FLOPs (speedup) between Ours and MaskPredict on each dataset.}
\label{flops_compare}
\end{table*}

\subsection{Sentence Length}
\begin{figure}[h!] 
  \begin{minipage}[b]{0.5\textwidth} 
    \centering 
    \includegraphics[width=0.8\textwidth]{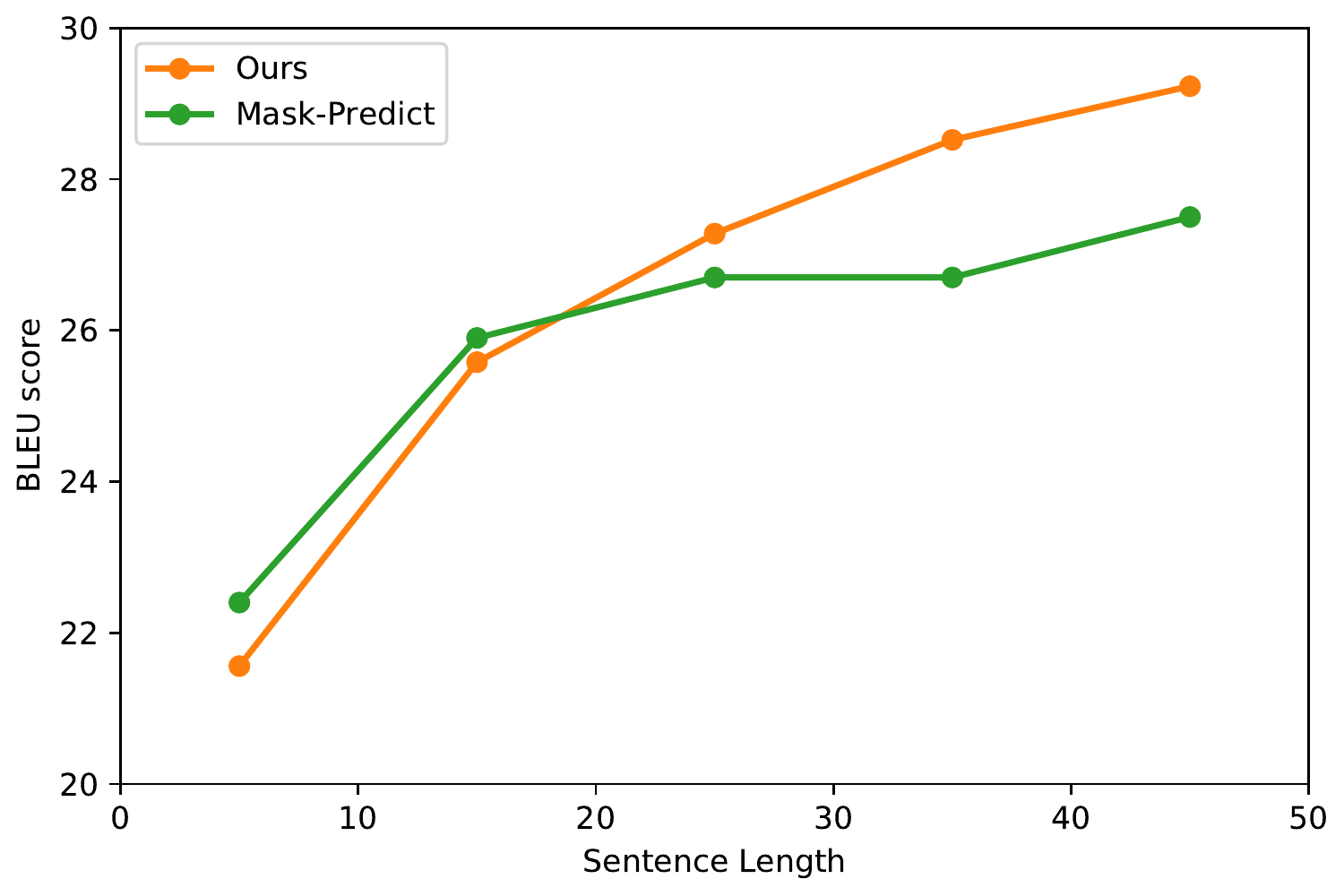} 
    \caption{The Effect of Sentence Length} 
    \label{sentence} 
  \end{minipage} 
  \begin{minipage}[b]{0.5\textwidth} 
    \centering
    \begin{tabular}{ccc} 
    \hline 
    \textbf{Iterations} & \multicolumn{2}{c}{\textbf{WMT14 EN\(\rightarrow\)DE}}\\
    & \textbf{MaskPredict} & \textbf{Ours}  \\
    \hline
    T = 1 & 16.72\% & 9.67\% \\
    T = 2 & 5.40\%  & 3.34\% \\
    T = 3 & 2.03\%  & 1.48\% \\
    T = 4 & 1.07\%  & 0.89\% \\
    T = 5 & 0.72\%  & 0.60\% \\
    \hline
    \end{tabular} 
    \captionof{table}{The percentage of repetitive words at different number of decoding iterations (T).} 
    \label{repeating_tokens} 
  \end{minipage} 
\end{figure}

Compared with Mask-Predict, we step further to examine the influence of Self-Review Mechanism on different sentence lengths. We conduct comparative experiments on WMT14 EN\(\rightarrow\)DE, and divide the reference target by length into different buckets. As shown in Figure~\ref{sentence}, Mask-Predict performs similar or slightly better than our model when the sentence length is small. However, the performance of our model is significantly improved as the sentence length increases, even leading to a wide gap with Mask-Predict when the sentence length is quite large. This result supports that our proposed Self-Review Mechanism is better at capturing the long-term dependency among the target words. 

\subsection{Adjacent Words}
\begin{figure*}[h!]
    \centering
    \subfigure[Mask-Predict]{
        \includegraphics[width=0.4\textwidth]{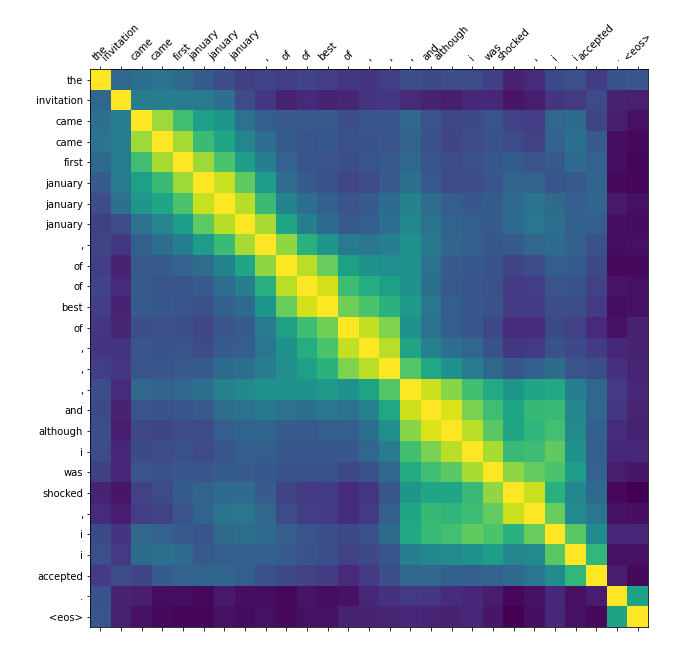}
    }
    \subfigure[Ours]{
	\includegraphics[width=0.44\textwidth]{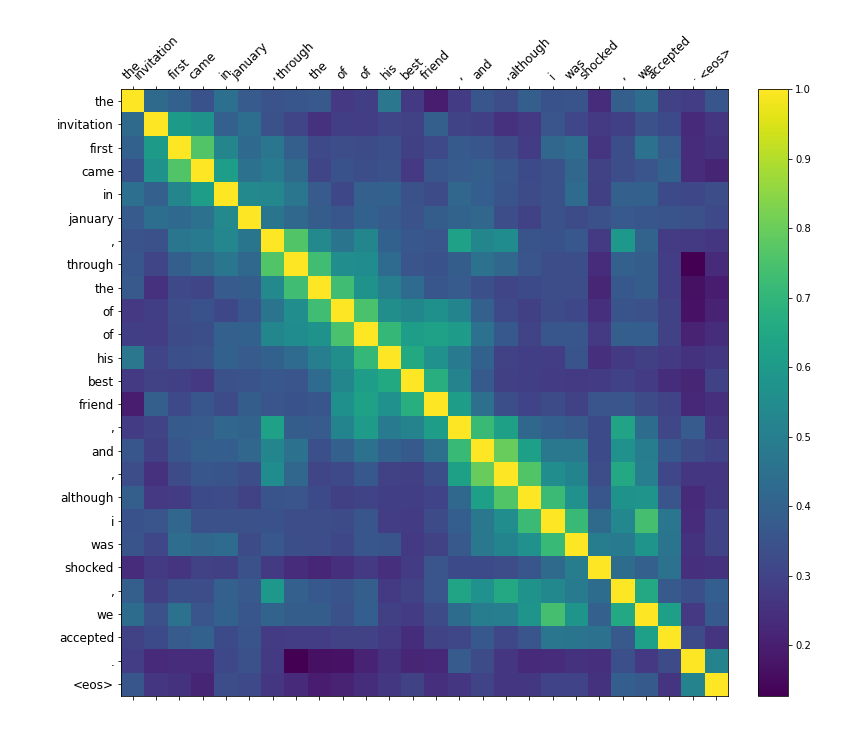}
    }
    \caption{The heatmaps of cosine similarities. The axis corresponds to the generated target word.}
    \label{cosine similarity}
\end{figure*}

According previous work~\cite{Wang2019NonAutoregressiveMT}, non-autoregressive models usually suffer from repetitive words at adjacent positions. To validate if such inappropriate pattern is remedied by Self-Review Mechanism, we conduct a statistical study of the repetitive words to compare Mask-Predict and our model. The results in Table~\ref{repeating_tokens} show that our model has substantially less repetitive words than Mask-Predict. For better understanding, we visualize the cosine similarities of two generated targets by Mask-Predict and our model respectively given a same input source, where the similarities are measured between decoder hidden states of the last layer. From the heatmaps of the resulting cosine similarities in Figure~\ref{cosine similarity}, we can see that there are observably more yellow blocks in (a) than those in (b), indicating that Mask-Predict shares much more similar hidden states across the positions of the generated sentence, especially illustrated along the diagonal parts in Figure~\ref{cosine similarity}. The results of Table~\ref{repeating_tokens} and Figure~\ref{cosine similarity} demonstrate that our proposed Self-Review Mechanism is beneficial for the model to reduce repetitive words, which further indicate that Self-Review is also an effective technique for CMTM to capture the strong correlations among the target words.

\section{Related Work}
To tackle the high latency of autoregression, many researchers attempt to use one-shot parallel decoding for machine translation. \newcite{Gu2018NonAutoregressiveNM} firstly proposed a non-autoregressive model by using fertilities as a latent variable. Later on, several works introduced different kinds of latent information to improve the performance. \newcite{Kaiser2018FastDI} used a sequence of discrete latent variables as the decoder inputs, \newcite{Li2019HintBasedTF} utilized the hints from the hidden states and word alignments of an autoregressive model, and \newcite{Ma2019FlowSeqNC} modeled a meaningful generative flow using latent variables. Although these methods are able to decode the target in one shot, they usually suffer from the cost of translation accuracy~\cite{ren2020study}. Worse still, they will never have a chance to remedy the flawed translation.

Our work resides in the research line of iterative parallel decoding. \newcite{lee2018deterministic} iteratively refined the generated outputs through a denoising autoencoder. However, the optimization is complicated, as they resort to a heuristic method of stochastic corruption on the training data.  Still along this line, our work is most relevant to \cite{ghazvininejad2019mask}, where they proposed a simple yet effective method, i.e., Mask-Predict decoding strategy. A major difference is that \newcite{ghazvininejad2019mask} resorts to a typical conditional masked translation model (CMTM), which is highly based upon the assumption of conditional independence. However, this assumption goes against the highly multimodal distribution of true target translations~\cite{Gu2018NonAutoregressiveNM}. To alleviate the issue, we develop a Self-Review Mechanism to infuse sequential information into the CMTM model.

We also get inspired by the idea of augmenting the model with a discriminative task~\cite{Clark2020ELECTRAPT}, in order to solve the computational inefficiency of CMTM. \newcite{Clark2020ELECTRAPT} introduced a discriminator (similar to our \textsc{ArDecoder}) that learns from all input words rather than a small masked subset. Then, they further finetuned the discriminator for the downstream tasks. The difference lies in that we throw out \textsc{ArDecoder} and only perform iterative decoding on \textsc{Encoder-Decoder}. Besides, we tie the weights of \textsc{ArDecoder} and \textsc{Decoder} to ensure that our \textsc{Decoder} can take advantage of the sequential information learned from the discriminative task.

\section{Conclusion}
In this paper, we identify the drawback of CMTM that it is insufficient to capture the sequential correlations among target words. To tackle it, we propose a Self-Review Mechanism that is able to infuse sequential information into CMTM. On several benchmark datasets, we demonstrate that our approach achieves a huge improvement against previous non-autoregressive models and a competitive result to the state-of-the-art Transformer model. Through ablation study,  our proposed mechanism is also proved to speed up the training of a CMTM model.

\bibliographystyle{coling}
\bibliography{coling2020}

\end{document}